\documentclass[runningheads]{llncs}
\usepackage[T1]{fontenc}
\usepackage{graphicx}
\usepackage{amsmath}
\usepackage{amsfonts}
\usepackage{subcaption}
\usepackage{hyperref}
\usepackage[table, dvipsnames]{xcolor}
\usepackage{arydshln}

\usepackage[misc]{ifsym} 

\begin{document}
\title{Generalized Robust Fundus Photography-based Vision Loss Estimation for High Myopia}
\titlerunning{Generalized Robust Fundus Photo-based Vision Loss Estimation for HM}

\author{
Zipei Yan\inst{1}
\and
Zhile Liang\inst{2}
\and
Zhengji Liu\inst{3}
\and
Shuai Wang\inst{5}
\and
Rachel Ka-Man Chun\inst{3,4}
\and
Jizhou Li\inst{1}
\and
Chea-su Kee\inst{3,4}
\and
Dong Liang\inst{3,4,\textrm{\Letter}}
}

\authorrunning{Z. Yan et al.}

\institute{
School of Data Science, City University of Hong Kong, Kowloon, Hong Kong
\and
School of Engineering and Technology, The University of Newcastle, Australia
\and
School of Optometry, The Hong Kong Polytechnic University, Kowloon, Hong Kong
\email{dong1.liang@connect.polyu.hk}
\and
Centre for Eye and Vision Research, 17W Hong Kong Science Park, Hong Kong
\and
Department of Biomedical Engineering, Tsinghua University, Beijing, China
}

\maketitle              

\begin{abstract}
High myopia significantly increases the risk of irreversible vision loss. Traditional perimetry-based visual field (VF) assessment provides systematic quantification of visual loss but it is subjective and time-consuming. Consequently, machine learning models utilizing fundus photographs to estimate VF have emerged as promising alternatives. However, due to the high variability and the limited availability of VF data, existing VF estimation models fail to generalize well, particularly when facing out-of-distribution data across diverse centers and populations. To tackle this challenge, we propose a novel, parameter-efficient framework to enhance the generalized robustness of VF estimation on both in- and out-of-distribution data. Specifically, we design a Refinement-by-Denoising (RED) module for feature refinement and adaptation from pretrained vision models, aiming to learn high-entropy feature representations and to mitigate the domain gap effectively and efficiently. Through independent validation on two distinct real-world datasets from separate centers, our method significantly outperforms existing approaches in RMSE, MAE and correlation coefficient for both internal and external validation. Our proposed framework benefits both in- and out-of-distribution VF estimation, offering significant clinical implications and potential utility in real-world ophthalmic practices.



\end{abstract}

\section{Introduction}

High myopia (HM) is a significant risk factor for irreversible vision impairment, primarily due to its association with several ocular conditions, such as myopic maculopathy and retinal detachment~\cite{ohno2021imipm}. The assessment of visual field (VF) sensitivity is crucial for evaluating the risk and extent of vision loss, as it provides a systematic quantification of visual function~\cite{phu2017VFvalue}. However, traditional VF assessment using perimetry is notably subjective and time-consuming, as it highly requires patients' compliance throughout the test~\cite{lewis1986VFlimit}. 

Unlike VF assessment, fundus photography as an imaging technique for retinal morphology, not only offers an objective measurement of the retinal structure, but also emerged as a valuable alternative for evaluating the retinal function, based on the theory of ``structure-function relationship''~\cite{xie2022strufunc}. Recent advances deep learning-based methods~\cite{lee2020estimating,christopher2020deep,yan2023vfhm} have shown potential in leveraging fundus photos for accurate vision loss estimation, even point-wise VF estimation~\cite{yan2023vfhm}. These methods provide more objective and efficient alternatives to assess the risk and extent of vision loss.

Despite the potential of deep learning-based VF estimations, their practical implementation in clinical settings is compromised by two primary challenges associated with VF data. Firstly, the inherent high variability within VF data—largely caused by its subjective assessment~\cite{wild1999vfvariability}—and the variation in imaging devices and protocols across diverse centers and populations, will introduce unforeseen shifts in data distribution. This phenomenon, known as out-of-distribution (OOD) data, hinders the models' ability to generalize across datasets, presenting a significant challenge to the robustness and reliability of VF estimation. Secondly, the limited available VF data—mainly due to the cost and difficulty of acquisition~\cite{delgado2002vfcost}—further poses a significant barrier to the development of a robust deep learning model for VF estimation.  

To tackle these challenges, our study proposes a novel parameter-efficient framework to enhance the generalized robustness of VF estimation on both in-distribution and OOD data. Specifically, we design a Refinement-by-Denoising (RED) module for feature refinement and adaptation from pretrained vision models, aiming to mitigate noises originating from the domain gap and learn high-entropy feature representations for VF estimation. 

In brief, our method utilizes pretrained vision models for initial feature extraction from fundus photos. While these models are originally trained on natural images, their application to fundus photos introduces a domain gap and leads to corrupted raw features, mainly due to the differences between the natural image and fundus image domains. Therefore, to mitigate this domain gap, RED refines the raw features by unsupervised denoising, which not only removes the noise but also encourages high-entropy representations, facilitating the downstream VF estimation. As a result, our method significantly outperforms existing approaches in the validation of two distinct real-world datasets.

Our contributions are summarized as follows:
\begin{itemize}
    \item We propose a novel parameter-efficiency framework, RED, for robust VF estimation from fundus photos. RED is designed for feature refinement and adaptation from pretrained vision models, and could learn high-entropy feature representations and mitigate the domain gap effectively. 
    \item Our proposed RED outperforms existing models in RMSE, MAE and correlation coefficient for both internal and external validation on two distinct real-world datasets, offering significant clinical implications and potential utility in future ophthalmic practices.  
    \item This is the first work to systematically assess the robustness of VF estimation models across datasets from different centers and populations. 
\end{itemize}

\section{Problem Formulation}

Given a training set $\mathcal{D}=\{(\boldsymbol{X}^i,\boldsymbol{y}^i) \}_{i=1}^{N}$, where $\boldsymbol{X}^i \in \mathbb{R}^{H\times W \times C}$ denotes the fundus photo, $\boldsymbol{y}^i \in \mathbb{R}^{M}$ denotes its corresponded vectorized VF. The objective is to train a model on the given $\mathcal{D}$, which learns a mapping from the fundus photo to the target VF. The challenges mainly come from the limited training set. The training set is in a limited size and the fundus photo is in a high-dimensional space, therefore learning the model is challenging. Besides, conventional regression fails to predict VF accurately~\cite{yan2023vfhm}, due to its inability to learn high entropy feature representations~\cite{zhang2023improving}. Furthermore, we aim to learn a robust model that is generalized to both in- and out-of-distribution data.

\section{Proposed Method}

\begin{figure}[t]
    \centering
    \includegraphics[width=0.9\textwidth]{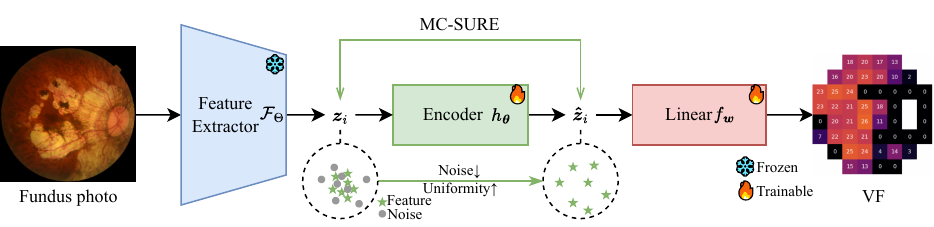}
    \caption{Overview of the proposed method.}
    \label{fig:overview}
\end{figure}

\subsection{Overview}

As illustrated in Fig.~\ref{fig:overview}, we present an overview of the proposed method. Specifically, we address this problem from a feature refinement perspective. Our methods consist of three components: feature extraction, feature denoising and regression. For feature extraction, we utilize pretrained vision models to extract the feature vectors from the fundus photo, thereby significantly reducing its dimensionality. However, the pretrained vision models are generally trained in the natural image domain; therefore, there is a domain gap between the natural image and the fundus photo domain. To address this problem, we propose a refinement by denoising (RED) module, which removes the noise in features and benefits the following regression process. Thereafter, the final regression learns to predict VF based on the refined features.

\subsection{Feature Refinement by Denoising (RED)}

Given a pretrained vision model $\mathcal{F}_{\Theta}(\cdot): \mathbb{R}^{H \times W \times C} \to \mathbb{R}^{K}$ where $\Theta$ denotes its parameters, for a fundus photo $\boldsymbol{X}$, its feature vector can be extracted as follows:
\begin{equation}
    \boldsymbol{z} = \mathcal{F}_{\Theta}(\boldsymbol{X}),
\end{equation}
where $\boldsymbol{z} \in \mathbb{R}^K$ denotes the extracted feature vector.

However, the pretrained vision models are generally trained in the natural image domain, and our input modality is the fundus photo, which is captured by the specialized digital camera. Although it is in RGB mode, there is a domain gap between the natural image domain and fundus photo domain. Following~\cite{li2022uncertainty}, we model the domain gap by assuming that each feature vector $\boldsymbol{z}$ is corrupted by additive white Gaussian noise (AWGN), which is formulated as follows:
\begin{equation}
    \boldsymbol{z} = \boldsymbol{z}^{\star} + \boldsymbol{n}, 
    \label{eq:noise}
\end{equation}
where $\boldsymbol{z}^{\star} \in \mathbb{R}^K$ denotes the underlying cleaned signal and $\boldsymbol{n} \in \mathbb{R}^K$ denotes an \textit{i.i.d.} Gaussian noise, i.e., $n_k \sim \mathcal{N}(0, \sigma^2)$.

\begin{table}[t]
    \centering
    \caption{Main results. $^\dag$ denotes fine-tuning the feature extractor.}
    \label{tab:main}
    \resizebox{\linewidth}{!}{
    \begin{tabular}{l|c|c:c:c|c:c:c}
        \hline
         & Trainable & \multicolumn{3}{c|}{Validation set} & \multicolumn{3}{c}{Test set} \\
        Method & Params (MB) & RMSE(\textdownarrow) & MAE(\textdownarrow) & PCC(\textuparrow) & RMSE(\textdownarrow) & MAE(\textdownarrow) & PCC(\textuparrow) \\
        \hline
        Classification & 1.04 & 4.96 & 3.81 & 0.32 & 3.17 & 2.39 & 0.19 \\
        EMD~\cite{hou2016squared} & 1.04 & 4.94 & 3.76 & 0.45 & 3.19 & 2.38 & 0.32 \\
        SOFT~\cite{bertinetto2020soft} & 1.04 & 4.95 & 3.80 & 0.33 & 3.16 & 2.39 & 0.20 \\
        CORAL~\cite{cao2020rank} & 1.02 & 4.78 & 3.68 & 0.41 & 3.14 & 2.43 &  0.30 \\
        OLL~\cite{castagnos2022oll} & 1.04 & 4.57 & 3.45 & 0.55 & 3.08 & 2.37 & 0.40 \\
        \hline
        VF-HM~\cite{yan2023vfhm} & 11.02$^\dag$ & 4.69 & 3.61 & 0.49 & 3.32 & 2.58 & 0.22  \\
        \hline
        Regression & 0.03 & 5.09 & 4.00 & 0.44 & 3.99 & 3.44 & -0.08 \\
        Regression + OE~\cite{zhang2023improving} & 0.53 & 4.36 & 3.25 & 0.60 & 3.08 & 2.39 & 0.31 \\
        \rowcolor{lightgray} Regression + \textbf{RED} & 0.53 & \textbf{4.21} & \textbf{3.13} & \textbf{0.63} & \textbf{2.92} & \textbf{2.19} & \textbf{0.53} \\
        \hline
    \end{tabular}
    }
\end{table}

Thereafter, we propose a Refinement by Denoising (RED) module for mitigating this gap by denoising. Specifically, we parameterize RED by an encoder $h_{\boldsymbol{\theta}}(\cdot): \mathbb{R}^{K} \to \mathbb{R}^{K}$, which denoises the features by $\boldsymbol{\hat{z}} = h_{\boldsymbol{\theta}}(\boldsymbol{z})$. However, the ground truth $\boldsymbol{z}^{\star}$ is not available in practice, therefore optimizing $h_{\boldsymbol{\theta}}(\cdot)$ with the mean squared error (MSE) is impossible. And the MSE is defined as follows:
\begin{equation}
    \text{MSE}(h_{\boldsymbol{\theta}}(\boldsymbol{z})) = \frac{1}{K} \Vert \boldsymbol{z}^{\star} - \boldsymbol{\hat{z}} \Vert^2 = \frac{1}{K} \Vert \boldsymbol{z}^{\star} - h_{\boldsymbol{\theta}}(\boldsymbol{z}) \Vert^2,
    \label{eq:mse}
\end{equation}
where $\Vert \cdot \Vert$ denotes $\ell^2$ norm.

Instead, following Stein's unbiased risk estimator (SURE)~\cite{stein1981sure}, we can derive the unbiased estimation of the above MSE in Eq.\eqref{eq:mse}, which only requires noisy feature $\boldsymbol{z}$. And the SURE is formulated as follows:

\begin{equation}
    \text{SURE}(h_{\boldsymbol{\theta}}(\boldsymbol{z})) = \frac{1}{K} \Vert \boldsymbol{z} - h_{\boldsymbol{\theta}}(\boldsymbol{z}) \Vert^2 - \sigma^2 + \frac{2\sigma^2}{K} \underbrace{ \sum_{k=1}^{K}\frac{\partial h_{\boldsymbol{\theta}}(\boldsymbol{z})_k }{\partial \boldsymbol{z}_k} }_{\text{divergence}}.
    \label{eq:sure}
\end{equation}

\begin{theorem}
    \label{thm:sure}
    (\cite{stein1981sure}) The random variable $\text{\normalfont SURE}(h_{\boldsymbol{\theta}}(\boldsymbol{z}))$ is an unbiased estimator of $\text{\normalfont MSE}(h_{\boldsymbol{\theta}}(\boldsymbol{z}))$, that is, $\mathbb{E}_{\boldsymbol{n}}\{\text{\normalfont MSE}(h_{\boldsymbol{\theta}}(\boldsymbol{z}))\} = \mathbb{E}_{\boldsymbol{n}}\{\text{\normalfont SURE}(h_{\boldsymbol{\theta}}(\boldsymbol{z}))\}$,
\end{theorem}
where $\mathbb{E}_{\boldsymbol{n}}\{\cdot\}$ represents the expectation with respect to $\boldsymbol{n}$.

For fast approximating the last divergence term in Eq.\eqref{eq:sure}, we follow the MC-SURE~\cite{ramani2008mcsure}, which utilizes Monte Carlo method~\cite{metropolis1949monte} in approximation.

\begin{theorem} (\cite{ramani2008mcsure}) Let $\boldsymbol{b} \sim \mathcal{N}(\boldsymbol{0}, \sigma^2\boldsymbol{I}) \in \mathbb{R}^K$ be a zero-mean i.i.d. normal vector, then, 
\begin{equation}
    \sum_{k=1}^{K}\frac{\partial h_{\boldsymbol{\theta}}(\boldsymbol{z})_k }{\partial \boldsymbol{z}_k} = \lim_{\epsilon \to 0} \mathbb{E}_{\boldsymbol{b}} \Bigl\{ \boldsymbol{b}^T \Bigl( \frac{ h_{\boldsymbol{\theta}}(\boldsymbol{z} + \epsilon\boldsymbol{b}) - h_{\boldsymbol{\theta}}(\boldsymbol{z}) }{\epsilon} \Bigr) \Bigr\},
    \label{eq:mc_sure_div}
\end{equation}
\label{thm:mc_sure}
\end{theorem}
where $\epsilon$ denotes a fixed small positive scalar, and $(\cdot)^T$ denotes the transpose operator.

\begin{table}[t]
    \caption{Results of abnormal cases on Test set.}
    \label{tab:abnormal}
    \centering
    \resizebox{0.85\linewidth}{!}{
    \begin{tabular}{l|c:c:c|c:c:c}
        \hline
        Category & \multicolumn{3}{c|}{GHT: Outside Normal Limits} & \multicolumn{3}{c}{PSD: $P\leq 0.5\%$} \\
        \hline
        Method & RMSE(\textdownarrow) & MAE(\textdownarrow) & PCC(\textuparrow) & RMSE(\textdownarrow) & MAE(\textdownarrow) & PCC(\textuparrow) \\
        \hline
        Classification & 4.32 & 3.83 & 0.28 & 5.53 & 3.06 & 0.13 \\
        EMD~\cite{hou2016squared} & 4.55 & 3.57 & 0.49 & 5.57 & 3.05 & 0.32 \\
        SOFT~\cite{bertinetto2020soft} & 4.33 & 3.40 & 0.27 & 5.51 & 3.03 & 0.15 \\
        CORAL~\cite{cao2020rank} & 4.11 & 3.25 & 0.45 & 5.45 & 3.01 & 0.23 \\
        OLL~\cite{castagnos2022oll} & 4.10 & 3.21 & 0.58 & 5.40 & 2.94 & 0.34 \\
        \hline
        VF-HM~\cite{yan2023vfhm} & 4.25 & 3.38 & 0.17 & 5.54 & 3.18 & 0.12 \\
        \hline
        Regression & 4.38 & 3.67 & -0.23 & 5.52 & 3.70 & 0.38 \\
        Regression + OE~\cite{zhang2023improving} & 4.24 & 3.31 & 0.58 & 5.40 & 2.88 & 0.31 \\
        \rowcolor{lightgray} Regression + \textbf{RED} & \textbf{3.95} & \textbf{3.05} & \textbf{0.62} & \textbf{5.15} & \textbf{2.72} & \textbf{0.47} \\
        \hline
    \end{tabular}
    }
\end{table}

Based on the above Theorem \ref{thm:mc_sure}, the last divergence term in Eq.\eqref{eq:sure} can be approximated as follows~\cite{ramani2008mcsure,soltanayev2018training}:
\begin{equation}
    \frac{1}{K}\sum_{k=1}^{K}\frac{\partial h_{\boldsymbol{\theta}}(\boldsymbol{z})_k }{\partial \boldsymbol{z}_k} \approx \frac{1}{\epsilon K}\boldsymbol{b}^T \bigl( h_{\boldsymbol{\theta}}(\boldsymbol{z} + \epsilon\boldsymbol{b} ) - h_{\boldsymbol{\theta}}(\boldsymbol{z}) \bigr).
\end{equation}

Therefore, the training objective function of $h_{\boldsymbol{\theta}}(\cdot)$ is formulated as follows:
\begin{equation}
    \text{MC-SURE}(h_{\boldsymbol{\theta}}(\boldsymbol{z})) = \frac{1}{K} \Vert \boldsymbol{z} - h_{\boldsymbol{\theta}}(\boldsymbol{z}) \Vert^2 - \sigma^2 + \frac{2\sigma^2}{\epsilon K}\boldsymbol{b}^T \Bigl( h_{\boldsymbol{\theta}}(\boldsymbol{z} + \epsilon\boldsymbol{b} ) - h_{\boldsymbol{\theta}}(\boldsymbol{z}) \Bigr).
    \label{eq:mc_sure}
\end{equation}

Finally, we utilize a regression module $f_{\boldsymbol{w}}(\cdot): \mathbb{R}^K \to \mathbb{R}^M$ for predicting VF based on the denoised features, that is,  $\boldsymbol{\hat{y}} = f_{\boldsymbol{w}}(\boldsymbol{\hat{z}})$. And the overall training objective is summarized as follows:
\begin{equation}
    \mathcal{L} = \frac{1}{N} \sum_{i=1}^N \left( \frac{1}{M} \Vert \boldsymbol{\hat{y}}^i - \boldsymbol{y}^i \Vert^2 + \lambda \cdot \text{MC-SURE}(h_{\boldsymbol{\theta}}(\boldsymbol{z}^i)) \right),
\end{equation}
where $\lambda \in \mathbb{R}_+$ is a trade-off hyper-parameter.

\subsection{Analysis of the role of noise} \label{sec:analysis}

To understand the role of noise in features, we conduct analyses on both noisy features $\boldsymbol{z}$ and denoised features $\boldsymbol{\hat{z}}$, respectively. For simplicity, we consider a linear regression model $f_{\boldsymbol{w}}(\cdot): \mathbb{R}^{K} \to \mathbb{R}$ for predicting a single VF point based on a given feature; and without loss of generality, we ignore its bias term, and it is defined as:
\begin{equation}
    f_{\boldsymbol{w}}(\boldsymbol{z}) = \boldsymbol{w}^T \boldsymbol{z},
\end{equation}
where $\boldsymbol{w} \in \mathbb{R}^{K}$ denotes its weight parameter.

\begin{figure}[t]
    \centering
    \includegraphics[width=0.8\textwidth]{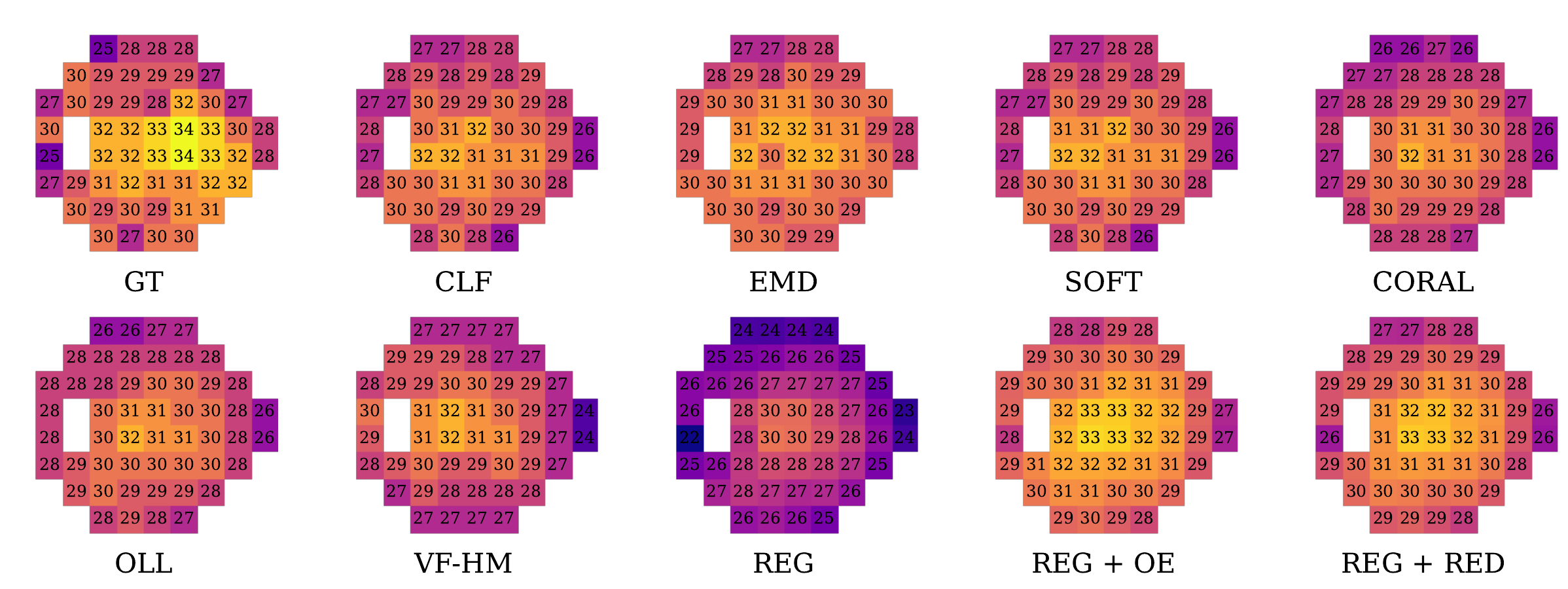}
    \caption{Visualization of predictions of a case in the Test set. }
    \label{fig:pred}
\end{figure}

Recall the noise formulation in Eq.\eqref{eq:noise} for features $\boldsymbol{z}$, and consider the training objective of $f_{\boldsymbol{w}}(\cdot)$ is a squared error, then it can be expressed as follows:
\begin{equation}
    \begin{aligned}
        (f_{\boldsymbol{w}}(\boldsymbol{z}) - y )^2 &=  \bigl( \boldsymbol{w}^T(\boldsymbol{z}^{\star} + \boldsymbol{n}) - y \bigr)^2 
        = \bigl( ( \boldsymbol{w}^T\boldsymbol{z}^{\star} - y ) + \boldsymbol{w}^T\boldsymbol{n} \bigr)^2 \\
        &= ( \boldsymbol{w}^T\boldsymbol{z}^{\star} - y )^2 + \Vert \boldsymbol{w}^T\boldsymbol{n} \Vert^2 \\
        &= ( \boldsymbol{w}^T\boldsymbol{z}^{\star} - y )^2 + \underbrace{K\sigma^2 \Vert \boldsymbol{w}\Vert^2}_{\text{weight decay}},
        \label{eq:wd_adaptive}
    \end{aligned}
\end{equation}
therefore, the noise in noisy features $\boldsymbol{z}$ can be interpreted as a regularization~\cite{chapelle2000vrm}, i.e., a constant weight decay.

Furthermore, we investigate the effect of denoised feature $\boldsymbol{\hat{z}}$. Inspired by  ~\cite{romano2015boosting,chang2018boost,chang2020boost,yan2024its}, we decompose it as follows:
\begin{equation}
    \boldsymbol{\hat{z}} = \boldsymbol{z}^{\star} + \boldsymbol{e},
\end{equation}
where $\boldsymbol{e}$ denotes the residual error, which is composed of the lost signals of $\boldsymbol{z}^{\star}$ and the remaining noise from $\boldsymbol{n}$.

Similarly, the training objective can be expanded as follows:
\begin{equation}
    (f_{\boldsymbol{w}}(\boldsymbol{\hat{z}}) - y )^2 = ( \boldsymbol{w}^T\boldsymbol{z}^{\star} - y )^2 + \underbrace{ \cos^2 (\boldsymbol{e}, \boldsymbol{w})  \Vert \boldsymbol{e}\Vert^2 \Vert \boldsymbol{w}\Vert^2}_{\text{weight decay}},
    \label{eq:wd}
\end{equation}
where the error in denoised features $\boldsymbol{\hat{z}}$ also performs as a regularization, yet, an adaptive weight decay depending on the residual error $\boldsymbol{e}$, as well as the quality of the denoising.

\section{Experiments}

\subsection{Experimental Data}
The experimental data (including training, validation, and test set) comes from the HM population. Specifically, the training and validation sets are from the same clinic site, while the test set is from another clinic site, there are no overlapping patients among them. The training set includes 254 eyes, the validation set includes 45 eyes,  and the test set consists of 92 eyes. Besides, the fundus photo is captured in RGB colorful mode, the VF is measured in the 24-2 mode with 52 effective points.

\subsection{Experimental Setup}

\textbf{Data pre-processing.} We choose the left eye pattern as our base~\cite{yan2023vfhm}, while we horizontally flip these data that are not in the left eye pattern.

\textbf{Baselines.} We mainly compare our method to classification and regression baselines, which utilize extracted features. Specifically, classification baselines including conventional and ordinal classifications, including EMD~\cite{hou2016squared}, SOFT~\cite{bertinetto2020soft}, CORAL~\cite{cao2020rank} and OLL~\cite{castagnos2022oll}. Besides, we also compare our method to VF-HM~\cite{yan2023vfhm}, which is based on CORAL and auxiliary learning. Additionally, we consider the regression regularized by ordinal entropy (OE)~\cite{zhang2023improving}.

\textbf{Evaluation methods.} Following~\cite{park2020deep,zheng2019glaucoma,xu2021pami,christopher2020deep,datta2021retinervenet,yan2023vfhm}, we utilize two metrics: RMSE and MAE for quantitative evaluation. In addition, we utilize the Pearson correlation coefficient (PCC) to measure the linear relationship between the predicted and ground-truth VF. For qualitative evaluation, we visualize some representative predictions from the test set.

\textbf{Implementation details.} We follow the official implementations for all baselines. And for a fair comparison, we utilize the ImageNet-1K pretrained ResNet-18~\cite{he2016deep} as the feature extractor. For the encoder $h_{\boldsymbol{\theta}}(\cdot)$ in RED and OE, we utilize a ReLU-based MLP with only one hidden layer\footnote{Our code is available at \url{https://github.com/yanzipei/VF_RED}}.

\begin{table}[t]
    \centering
    \begin{minipage}{.49\linewidth}
        \caption{Alternative denoising kernels.}
        \label{tab:denoise}
        \resizebox{1.0\linewidth}{!}{
        \begin{tabular}{l|c:c:c|c:c:c}
        \hline
        & \multicolumn{3}{c|}{Validation set} & \multicolumn{3}{c}{Test set} \\
        \hline
        Method & RMSE(\textdownarrow) & MAE(\textdownarrow) & PCC(\textuparrow) & RMSE(\textdownarrow) & MAE(\textdownarrow) & PCC(\textuparrow) \\
        \hline
        Regression & 5.09 & 4.00 & 0.44 & 3.99 & 3.44 & -0.08 \\
        \hline
        + Mean (k=2) & 5.01 & 3.93 & 0.14 & 3.65 & 3.05 & -0.10 \\
        + Mean (k=3) & 5.00 & 3.93 & 0.00 & 3.64 & 3.03 & -0.11 \\
        + Mean (k=4) & 5.02 & 3.94 & -0.05 & 3.63 & 3.03 & -0.11 \\
        \hdashline
        + Median (k=3) & 4.98 & 3.89 & 0.15 & 3.65 & 3.05 & -0.01 \\
        + Median (k=5) & 5.03 & 3.94 & -0.04 & 3.64 & 3.03 & -0.12 \\
        + Median (k=7) & 5.07 & 3.99 & -0.20 & 3.69 & 3.09 &  -0.14 \\
        \hline
        \rowcolor{lightgray} + \textbf{RED} & \textbf{4.21} & \textbf{3.13} & \textbf{0.63} & \textbf{2.92} & \textbf{2.19} & \textbf{0.53} \\
        \hline
        \end{tabular}
        }
    \end{minipage}
    \begin{minipage}{.49\linewidth}
        \caption{Alternative feature extractors.}
        \label{tab:feature}
        \resizebox{1.0\linewidth}{!}{
        \begin{tabular}{l|c:c:c|c:c:c}
        \hline
        & \multicolumn{3}{c|}{Validation set} & \multicolumn{3}{c}{Test set} \\
        \hline
        Method & RMSE(\textdownarrow) & MAE(\textdownarrow) & PCC(\textuparrow) & RMSE(\textdownarrow) & MAE(\textdownarrow) & PCC(\textuparrow) \\
        \hline
        ResNet-18~\cite{he2016deep} & 5.09 & 4.00 & 0.44 & 3.99 & 3.44 & -0.08 \\
        + RED & 4.21 & 3.13 & 0.63 & 2.92 & 2.19 & 0.53 \\
        \hline
        ConvNeXt-T~\cite{Liu2022CVPR} & 4.91 & 3.82 & 0.53 & 3.90 & 3.31 & -0.08 \\
        + RED & 4.39 & 3.26 & 0.57 & 3.01 & 2.27 & 0.49 \\
        \hline
        ViT-B-16~\cite{dosovitskiy2021image} & 4.93 & 3.84 & 0.31 & 4.10 & 3.58 & 0.06 \\
        + RED & 4.42 & 3.30 &  0.59 & 3.05 & 2.38 & 0.35 \\
        \hline
        Swin-T~\cite{Liu2021ICCV} & 4.85 & 3.75 & 0.46 & 4.18 & 3.64 & -0.12 \\
        + RED & 4.12 & 3.03 & 0.63 & 3.56 & 2.99 & 0.38 \\
        \hline
        \end{tabular}
        }
    \end{minipage}
\end{table}

\subsection{Experimental Results}

\textbf{Main results.} Table~\ref{tab:main} reports the results of different methods on both validation and test sets. In general, RED outperforms others on both validation and test sets. Specifically, for classifications, we observe that OLL achieves better performance and CORAL is the runner-up. Besides, VF-HM outperforms OLL on the validation set but is worse on the test set, indicating fine-tuning overfits and poorly generalizes to unseen OOD test data. For regressions, we observe that RED can improve the generalization on both in- and out-of-distribution data, according to the significant improvement from RMSE, MAE and PCC on both validation and test sets; meanwhile, OE mainly improves the generalization of in-distribution data.

\textbf{Results on abnormal cases.} Table~\ref{tab:abnormal} reports the results of abnormal cases on the Test set. Abnormal cases are identified based on Glaucoma Hemifield Test (GHT, outside normal limited) and Pattern Standard Deviation (PSD, $P \leq 0.5\%$). Similar to the main results, RED achieves better performance than others in abnormal cases, suggesting the RED's ability to identify and predict abnormal cases more accurately.

\textbf{Visualization of predictions.} Fig.~\ref{fig:pred} visualizes predictions from different models for a representative case on test data. Specifically, conventional regression fails to predict vision loss, while RED predicts local vision loss more precisely.

\subsection{Ablation Study}
Apart from the analysis in Sec~\ref{sec:analysis}, we conduct experiments to analyze the effectiveness of the proposed RED.

\begin{figure}[t]
    \centering
    \includegraphics[width=0.9\linewidth]{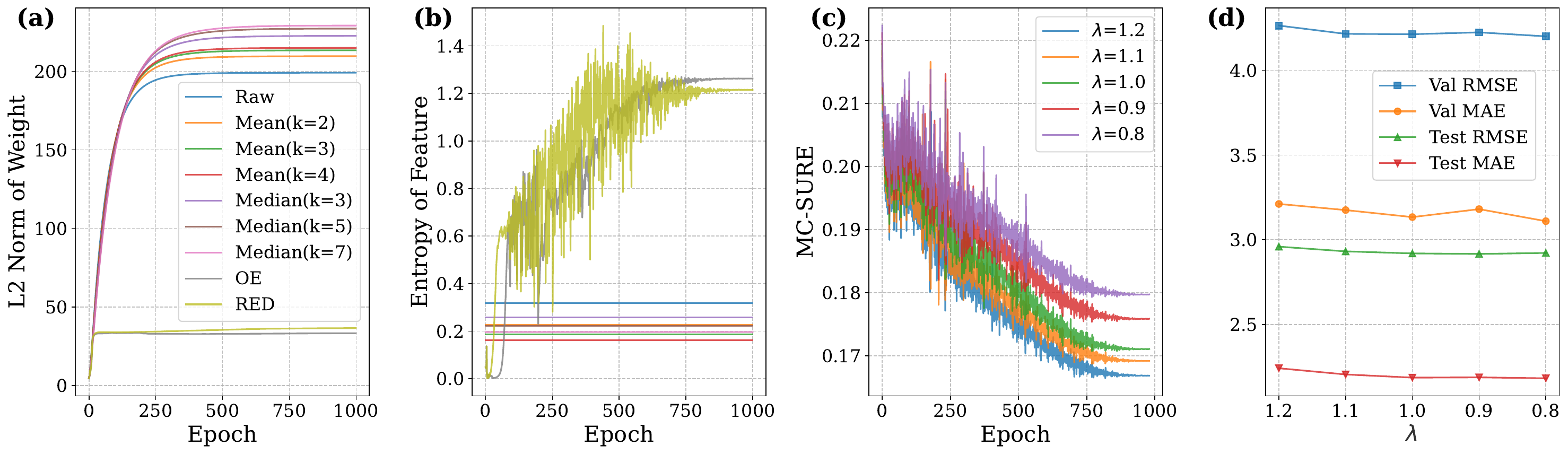}
    \caption{(a) L2 Norm of the weight of the regression layer (b) Entropy of feature space (c) MC-SURE loss during training, and (d) RMSE and MAE metrics corresponding to different $\lambda$.}
    \label{fig:ablation}
\end{figure}

\textbf{Effectiveness of RED.} We compared RED to different denoising kernels including mean and median kernels. In Table~\ref{tab:denoise}, we observe that RED is more effective than others. Besides, in Fig.~\ref{fig:ablation}(a) and (b) verify our method works as a weight decay regularized to smaller weights, and encourages to learn high entropy feature representations.

\textbf{Impact of Hyper-parameter.} As shown in Fig.~\ref{fig:ablation}(c), we explore the impact of different $\lambda$ for MC-SURE loss; besides, according to the result in Fig.~\ref{fig:ablation}(d), we observe that $\lambda =1.0$ is a good trade-off parameter. 

\textbf{Alternative feature extractors.} We also explore alternative feature extractors, such as ConvNeXT~\cite{Liu2022CVPR}, ViT~\cite{dosovitskiy2021image}, and Swin~\cite{Liu2021ICCV}. In Table~\ref {tab:feature}, we observe that RED has broad applicability to different feature extractors.

\section{Conclusion}
In this work, we propose a parameter-efficiency framework: RED for robust VF estimation from fundus photos and validate it on two distinct real-world datasets from separate centers. RED significantly outperforms existing methods in both internal and external validation, suggesting a more robust and reliable VF estimation approach. Our work benefits not only in-distribution but also out-of-distribution data, highlighting significant clinical implications and potential utility in future ophthalmic practices. A notable limitation of our work is the demographic homogeneity of the datasets used, which consisted exclusively of patients with non-glaucomatous HM from the same ethnicity, although they were collected from diverse sites using diverse imaging devices and protocols. This may affect the applicability of our model to a broader population with diverse demographics (e.g., ethnicity, ocular diseases). Further study is required to evaluate the influence of demographic background on VF estimation. 

\subsubsection{Acknowledgements} 
This work is supported by Centre for Eye and Vision Research (CEVR), InnoHK CEVR Project 1.5, 17W Hong Kong Science Park, Hong Kong; Research Centre for SHARP Vision (RCSV) and Research Institute for Artificial Intelligence of Things (RIAIoT), The Hong Kong Polytechnic University; and PolyU Grants (P0035514). 

\subsubsection{Disclosure of Interests}
The authors have no competing interests to declare that are relevant to the content of this article.

%
%

\bibliographystyle{splncs04}
\bibliography{ref}

\end{document}